\documentclass[letterpaper, 10 pt, conference]{ieeeconf}  

\IEEEoverridecommandlockouts                              

\usepackage{graphicx}
\usepackage{multicol}
\usepackage{footmisc}

\usepackage{amsthm}
\usepackage{amssymb}  
\usepackage{amsmath} 
\usepackage[font={small}]{caption}
\usepackage{url}
\usepackage{subcaption}
\usepackage{algorithmic}
\usepackage{algorithm}
\usepackage{xcolor}
\usepackage{eqparbox}
\usepackage{xspace}
\usepackage{lastpage}
\usepackage[T1]{fontenc}

\DeclareCaptionLabelFormat{andtable}{#1~#2  \&  \tablename~\thetable}

\newtheorem{problem}{Problem}
\renewcommand\vec{\mathbf}

\floatname{algorithm}{}

\newcommand{\x}{\mathcal X}
\newcommand{\z}{\mathcal Z}
\newcommand{\uspace}{\mathcal U}
\newcommand{\xfree}{\mathcal X_{\text{free}}}
\newcommand{\xobs}{\mathcal X_{\text{obs}}}
\newcommand{\xgoal}{\mathcal X_{\text{goal}}}
\newcommand{\xinit}{\vec{x}_{\text{init}}}
\newcommand{\zinit}{\vec{z}_{\text{init}}}
\newcommand{\zgoalregion}{\mathcal Z_{\text{goal}}}
\newcommand{\zgoal}{\vec{z}_{\text{goal}}}
\newcommand{\reals}{\mathbb{R}}

\newcommand{\algname}{L-SBMP\xspace}
\newcommand{\algtree}{L2RRT\xspace}

\newcommand{\enc}{h^\text{enc}_\phi}
\newcommand{\dec}{h^\text{dec}_\theta}
\newcommand{\dyn}{h^\text{dyn}_\psi}
\newcommand{\cc}{h^\text{cc}_\zeta}
\newcommand{\dx}{d_x}
\newcommand{\dz}{d_z}
\newcommand{\du}{d_u}

\newcommand{\rrt}{RRT\xspace}
\newcommand{\fmt}{FMT$^\ast$\xspace}

\graphicspath{{./fig/}}

\renewcommand{\baselinestretch}{0.98}

\title{\LARGE \bf
Robot Motion Planning in Learned Latent Spaces
}

\author{Brian Ichter$^{1}$ and Marco Pavone$^{1}$
\thanks{*This work was supported by NASA under the Space Technology Research Grants Program, Grant NNX12AQ43G, and by the King Abdulaziz City for Science and Technology (KACST). The GPUs used for this research were donated by the NVIDIA Corporation.}
\thanks{$^{1}$Department of Aeronautics and Astronautics, Stanford University, Stanford, CA 94305, USA
        {\tt\small \{ichter,pavone\}@stanford.edu}}}

\begin{document}

\maketitle
\thispagestyle{empty}
\pagestyle{empty}

\begin{abstract}
This paper presents Latent Sampling-based Motion Planning (\algname), a methodology towards computing motion plans for complex robotic systems by learning a plannable latent representation.
Recent works in control of robotic systems have effectively leveraged local, low-dimensional embeddings of high-dimensional dynamics. 
In this paper we combine these recent advances with techniques from sampling-based motion planning (SBMP) in order to design a methodology capable of planning for high-dimensional robotic systems beyond the reach of traditional approaches (e.g., humanoids, or even systems where planning occurs in the visual space). 
Specifically, the learned latent space is constructed through an autoencoding network, a dynamics network, and a collision checking network, which mirror the three main algorithmic primitives of SBMP, namely state sampling, local steering, and collision checking. 
Notably, these networks can be trained through only raw data of the system's states and actions along with a supervising collision checker.
Building upon these networks, an RRT-based algorithm is used to plan motions {\em directly in the latent space} -- we refer to this exploration algorithm as Learned Latent RRT (\algtree).
This algorithm globally explores the latent space and is capable of generalizing to new environments. The overall methodology is demonstrated on two planning problems, namely a visual planning problem, whereby planning happens in the visual (pixel) space, and a humanoid robot planning problem.

\end{abstract}


\section{Introduction}

Motion planning is a fundamental problem in robotics whereby one seeks to compute a dynamically-feasible trajectory connecting an initial state to a goal region while avoiding obstacles.
This problem quickly increases in difficulty with the dimensionality of the system and the complexity of the dynamics \cite{LaValle2006}.
For some high-dimensional systems, and even differentially-constrained systems, sampling-based motion planning (SBMP) techniques have emerged as a particularly successful approach. 
SBMP avoids the explicit construction of the state space and instead maintains an implicit representation of it through three algorithmic primitives: sampling the state space, steering (or connecting) to nearby sampled states, and collision checking such connections \cite{LaValle2006,LaValleKuffner2001}.
This implicit representation is global and increasingly accurate as samples are added  \cite{LaValle2006}.
However, applying SBMP to robotic systems with, say, more than ten dimensions, quickly becomes intractable.
This precludes its application (at least directly) to cases where, for example, plans are computed from visual inputs \cite{watter2015embed,finn2017deep} or the dynamics are very high-dimensional and complex, as for humanoids \cite{brockman2016openai}.

A promising solution to address this challenge is represented by recent works on learning-based control, which 
leverage learned, low-dimensional manifolds embedded within the state space to which the dynamics of a robotic system are mainly restricted to \cite{watter2015embed,banijamali2017robust,finn2016deep,srinivas2018universal}.
For example, a bipedal humanoid walking robot is generally in gait and upright, and a visual representation of a robot has a ``true" underlying lower-dimensional dynamical representation. These works, however, primarily focus only in
local regions, and can not be directly applied to compute motion plans whereby one needs to {\em globally} explore the state space. Our key insight to scale these techniques to robot motion planning is to leverage the local effectiveness of low-dimensional latent representations to learn a latent space in which one can directly use SBMP. In other words, the idea is to build a {\em global} implicit representation of the latent space through only local connections.
The latent space is constructed through an autoencoding network, a dynamics network, and a collision checking network, which mirror the three aforementioned algorithmic primitives of SBMP, namely state sampling, local steering, and collision checking. Specifically, the space is learned through an autoencoding network with learned latent dynamics, which provides
the ability to sample the latent space and connect local states through dynamically-feasible trajectories.
Notably, this network only requires raw data of the system's states and control inputs to be trained.
Separately, a collision checking network is trained to predict whether a trajectory between two latent states is collision free given a specific problem environment. Collectively, these networks provide a methodology to globally search the latent space (in our case, through an RRT-based algorithm), in a way that generalizes well to new environments and is general enough to apply to a broad class of problems, including visual planning and complex dynamics problems.
The final trajectory can then be projected to the full state space through the autoencoding network and used in conjunction with the aforementioned local control methods to effectively control the robotic system along the planned trajectory. We refer to this methodology as Latent Sampling-based Motion Planning (\algname).

{\em Related Work:} The field of representation learning seeks to learn representations of data to extract useful information for classifiers or other predictors \cite{bengio2013representation,roweis2000nonlinear,hinton2006reducing}.
This often takes the form of manifold learning, i.e., learning lower-dimensional manifolds embedded within high-dimensional data. 
Representation learning has been applied along the entire robotic stack -- from perception \cite{piater2011learning}, to decision making and planning \cite{co2018self,SchmerlingLeungEtAl2018}, to control and dynamics \cite{watter2015embed,banijamali2017robust,finn2016deep,chen2016dynamic}. Recently, these methods have seen particular success in the field of control. Embed to Control \cite{watter2015embed} and Robust Locally-Linear Controllable Embeddings \cite{banijamali2017robust} present approaches to learning to control robotic systems directly from raw images. 
Both methods learn a locally-linear embedding via a variational autoencoder and directly perform stochastic optimal control on the latent space. 
\cite{finn2017deep} learns to predict the results of actions in visual spaces and uses model-predictive control to plan local manipulator actions. In contrast, our work learns a {\em non-linear} latent space (allowing generalization to more complex systems) and a collision checking network (allowing generalization to new environments).

In the context of robot motion planning, Universal Planning Networks (UPN) \cite{srinivas2018universal} learn a plannable latent representation in which gradient descent can be used to compute a plan that minimizes an imitation loss (learned from an expert planner).
Our methodology, \algname, learns a similar latent space to UPN (by enforcing dynamics in the latent space), but performs a {\em global} search of the latent space to find motion plans.
In addition, \algname does not rely on an expert planner, which for many problems may not be available.
\cite{ha2017high} uses a Gaussian process latent variable model to learn a  low-dimensional representation and then uses inference to solve the planning problem via a particle-based approach. Compared to \algname, the latent variable model must be well initialized and using inference does not allow a global search of the latent space.
Motion Planning Networks (MPNet) \cite{qureshi2018motion} learns a low-dimensional representation of the obstacle space and uses a feedforward neural network to predict the next step an optimal planner would take given an initial state, goal state, and the obstacle representation (trained via an asymptotically optimal planner).
MPNet approaches the problem of accelerating solutions to planning problems for which solutions to similar problems can be provided by demonstration, whereas our work attempts to solve previously intractable planning problems.
Lastly, within sampling-based planning, learning has been used to replace parts of the algorithmic stack, such as the collision checker \cite{das2017fastron} or the sample distribution \cite{IchterHarrisonEtAl2018}. 
\algname too learns a collision checking network, though directly in the latent space, and, in a manner, learns where to sample by choosing samples for the exploration algorithm directly from encoded training states.

It is important to note that, in practice, the \algname methodology can be, and should be, combined with many of the aforementioned approaches. 
In many cases, \algname may be an initialization for the latent spaces in \cite{watter2015embed,banijamali2017robust,srinivas2018universal}. 
Furthermore, these approaches can be used to improve the trajectory generated by \algname and to control the motion along it.

{\em Statement of Contributions:} The contribution of this paper is threefold. 
First, we present the  Latent Sampling-based Motion Planning (\algname) methodology, which comprises an autoencoding network, a dynamics network, and a collision checking network to enable planning through the latent space. 
These networks can be trained through only raw data of the system's states and control inputs along with a supervising collision checker.
Second, building upon \algname, we present an RRT-based algorithm,  termed Learned Latent Rapidly-exploring Random Trees (\algtree), to plan motions {\em directly in the latent space}. \algtree efficiently and globally explores the learned manifold by maintaining an implicit representation of the space formed through sampling encoded states and dynamically propagating nearby trajectories in the tree (under the supervision of the collision checking network). 
The final latent trajectory is then projected back to the full state space through the autoencoding network. Third, we demonstrate our approach on two planning problems that are well beyond the reach of standard SBMP, namely a visual planning problem, whereby planning happens in the visual (pixel) space, and a humanoid robot planning problem.

{\em Organization:}  The remainder of the paper is organized as follows. Section \ref{sec:problemStatement} introduces the problem we are approaching and provides necessary background. Section \ref{sec:latentSpace} outlines the \algname methodology for learning the plannable latent space. Section \ref{sec:planning} presents \algtree, the sampling-based exploration strategy used to globally search the latent space. Section \ref{sec:experiments} demonstrates the performance of \algname on a visual planning problem and a humanoid planning problem. Section \ref{sec:conclusions} summarizes our findings and suggests several avenues for future work.

\section{Problem Statement}\label{sec:problemStatement}

In this section we formulate the motion planning problem we wish to solve.
We then briefly overview sampling-based motion planning algorithms in order to give necessary background and to motivate our methodology.
Lastly, we present the problem of learning plannable latent spaces.

\subsection{Robot Motion Planning Problem}

Let $\x \subseteq \reals^{\dx}$ and $\uspace \subseteq \reals^{\du}$ denote the state space and control input space, respectively, of a robotic system.
Let the dynamics of the robot be defined by 
\begin{equation}
\vec{\dot{x}}(\tau) = f^{(\text{c})}_\x(\vec{x}(\tau),\vec{u}(\tau)),
\end{equation}
where $\vec{x}(\tau) \in \x$ and $\vec{u}(\tau) \in \uspace$ denote the state and control input of the system at time $\tau$ and $f^{(\text{c})}_\x$ denotes the continuous-time system dynamics. 
To make the dynamics learning problem tractable we consider a discrete-time formulation.
Note however that we still consider the continuous-time underlying dynamics to determine collisions.
Let the discrete-time dynamics of the robot be defined by  
\begin{equation}
\vec{x}_{t+1} = f_\x(\vec{x}_t,\vec{u}_t),
\end{equation}
where $\vec{x}_t \in \x$ and $\vec{u}_t \in \uspace$ denote the state and control input of the system at time step $t$ and $f_\x$ denotes the discrete-time system dynamics. 

Let $\xfree \subseteq \x$ define the free state space of the robot and let $\xobs \subset \x$ define the obstacle space, such that $\xfree = \x \setminus \xobs$.
Let us be given an initial state $\xinit \in \xfree$ and a goal region $\xgoal \subset \xfree$.
A trajectory $\pi$ for the robot is defined as a series of states and control inputs $(\vec{x}_0,\vec{u}_0,\ldots,\vec{x}_T,\vec{u}_T)$, where $T$ is the number of time steps in the trajectory.
Further, let $\overline{\vec{x}_0,\ldots,\vec{x}_T}$ denote the continuous curve traced by the robot's trajectory (connecting $\vec{x}_0,\ldots,\vec{x}_T$).
We make this distinction here to clarify that collision checking should be done along the continuous trajectory connecting discrete states.
A trajectory is said to be \emph{collision-free} if $\overline{\vec{x}_0,\ldots,\vec{x}_T} \cap \xobs = \emptyset$.
A trajectory is said to be \emph{feasible} if it connects the initial state and goal region, i.e., $\vec{x}_0 = \xinit$ and $\vec{x}_T \in \xgoal$, is collision-free, and is dynamically feasible, i.e., $\vec{x}_t = f_\x(\vec{x}_{t-1},\vec{u}_{t-1})$ for $t = 1,\ldots,T$. In this work, we are interested in the following problem,

\begin{problem}[Motion Planning Problem]\label{prob:mp}
Given a motion planning problem $(\xinit, \xgoal, \xfree)$, find a feasible trajectory $\pi$. If no such trajectory exists, report failure.
\end{problem}

\subsection{Sampling-based Motion Planning}

Sampling-based motion planning has emerged as a highly effective algorithmic paradigm for solving complex motion planning problems.
SBMP avoids the explicit construction of the state space and instead maintains an implicit, searchable representation. 
This representation is formed by a set of probing samples which are locally connected and verified via a ``black box'' collision checking module \cite{LaValle2006}.
This methodology enables a global search and guarantees on probabilistic completeness and often asymptotic optimality \cite{KaramanFrazzoli2011,JansonSchmerlingEtAl2015}. 

One particularly successful family of sampling-based motion planning algorithms is the rapidly-exploring random tree (RRT) \cite{LaValleKuffner2001}. 
As our exploration algorithm is built from this, we now overview RRT to clarify each algorithmic primitive.
The algorithm begins by initializing a tree $\mathcal{T}$ with the initial state $\xinit$, and subsequently iteratively builds a tree through the state space.
At each iteration, a state in the free state space is sampled, $\vec{x}_\text{sample}\in\xfree$. 
The nearest node within the tree to $\vec{x}_\text{sample}$ is selected as the expansion node, $\vec{x}_\text{expand}$.
The node is then expanded towards $\vec{x}_\text{sample}$, forming a new sample $\vec{x}_\text{new}$.
If the trajectory connecting $\vec{x}_\text{expand}$ and $\vec{x}_\text{new}$ does not intersect $\xobs$, i.e., is collision-free, then $\vec{x}_\text{new}$ is added as a node to the tree.
This continues until either time elapses or a pre-determined number of samples is reached.
The approach is capable of globally searching the state space only through three algorithmic primitives: (1) sampling, (2) locally connecting, and (3) collision checking.
The key idea in this work is that, by ensuring each of these algorithmic primitives can be applied in a learned, low-dimensional latent space representing the robotic system, we can directly use SBMP to compute plans for the full state space.

\subsection{Plannable Latent Spaces}

The robotic systems we consider in this work may have complex-dynamics and high-dimensional state spaces. 
For example, these may be pixel representations of the robot and environment or they may be highly-dynamic, high-dimensional robots. 
In these cases standard approaches to solving motion planning problems are computationally intractable. 
Even approaches like SBMPs, which were initially created for high-dimensional systems, have difficulty scaling well beyond ten dimensions even in the absence of dynamic constraints. 

Our goal is to find the underlying manifold of the problem by learning a low-dimensional latent space model in which SBMP can be performed.
Due to the complexity of the system, this space should be learned only through sequences of states and control inputs of the robot in operation and a separate local collision checker.
Let $\z \subseteq \reals^{\dz}$ denote the learned latent space and $\vec{z}_t \in \z$ a latent state at time step $t$.
We seek to learn a mapping $m$ from $\x$ to $\z$ and a mapping $n$ from $\z$ to $\x$ for which latent dynamics $f_\z$ are enforced over $\z$.
We further learn a ``black-box'' collision checking function $g$ in the latent space. 
We thus wish to learn the following:
\begin{equation}
\begin{aligned}
&m : \x \to \z, \quad n : \z \to \x, \quad \vec{z}_{t+1} = f_\z(\vec{z}_t,\vec{u}_t), \\
&g(\vec{z}_t,\vec{z}_{t+1}, \xobs) = \{1 \text{ if } \overline{\vec{x}_t,\vec{x}_{t+1}} \cap \xobs = \emptyset, \, 0 \text{ otherwise} \}.
\end{aligned}
\end{equation}

\section{Learning a Latent Space for Robotic Motion Planning}\label{sec:latentSpace}

In order to plan on a low-dimensional latent space model, one needs the properties that: (1) the latent space can be sampled, (2) the dynamics are known, so as to connect nearby latent states, and (3) a local, latent trajectory can be classified as collision-free.
It is also desirable that the latent samples can be mapped to the full space once a solution is found, so that the robotic system can execute the motion plan.
Finally, we do not assume we have access to the true system dynamics $f_\x$, i.e., only observations of states and actions over time and a collision check.

The Latent Sampling-based Motion Planning (\algname) architecture consists of three networks: (1) an autoencoder, (2) a latent local dynamics model, and (3) a collision classifier. 
The autoencoder projects the high-dimensional full state ($\vec{x}_t$) into a lower-dimensional latent state ($\vec{z}_t$), 
and also allows decoding to form inputs for a subsequent controller.
The dynamics model enforces local dynamics on the encoded latent space and allows the dynamics to be propagated.
Finally, the collision classifier identifies if the connection between two nearby latent states will be possible given the problem environment. 
The latent space is trained from a series of trajectories from the robotic system in operation; i.e., a sequence of states $\vec{x}_t$ and actions $\vec{u}_t$.
The collision checking network is trained from a data set of nearby states and a label denoting if the connecting trajectory is in collision.

\subsection{Learned Latent Space and Dynamics}

The first network in our architecture follows a standard treatment for autoencoders. Given a robotic system with state $\vec{x}_t$ and latent state $\vec{z}_t$, we define an encoding network $\enc(\vec{x}_t)$ and a decoding network $\dec(\vec{z}_t)$, with parameters $\phi$ and $\theta$, respectively. 
These networks project the full state into the lower-dimensional latent space and then reconstruct the full state.
Depending on the robotic system, such networks may be convolutional or fully connected networks.
For this work we use an $\ell_2$ reconstruction loss.

The second network in our architecture learns, and subsequently enforces, local dynamics in the latent space. 
Let $\dyn(\vec{z}_t,\vec{u}_t)$ denote the dynamics network and $\psi$ its parameters. This network predicts the next state in the latent space, $\hat{\vec{z}}_{t+1}$.
The accuracy of this prediction is enforced through two losses. 
First, the prediction is decoded $\dec(\hat{\vec{z}}_{t+1})$, and compared to the full state $\vec{x}_{t+1}$ with an $\ell_2$ reconstruction loss.
Second, the prediction is compared in the latent space to the next encoded state $\vec{z}_{t + 1} = \enc(\vec{x}_{t+1})$.
In order to have a dynamically-consistent loss, we linearize the dynamics around $\vec{z}_t$ and $\vec{u}_t$ and compute the weighted controllability Gramian $G_t$ (also known as the reachability Gramian) as 
\begin{equation}\label{eq:linearization}
A_t = \frac{\partial \dyn}{\partial \vec{z}}(\vec{z}_t, \vec{u}_t), \, B_t = \frac{\partial \dyn}{\partial \vec{u}}(\vec{z}_t, \vec{u}_t), \, G_t = A_t B_t B_t^\top A_t^\top.
\end{equation}
The loss is then computed as the minimum energy between the prediction and the true next latent state, that is $\mathcal{L}_t^\text{dyn} = (\vec{z}_{t + 1} - \hat{\vec{z}}_{t + 1})^\top{G_t^{-1}}(\vec{z}_{t + 1} - \hat{\vec{z}}_{t + 1})$.
Essentially this enforces that the loss in prediction is representative of the dynamics of the system, e.g., for a car, missing the latent state prediction forward or backward is less important than missing left and right. 
Note however, that initially, before the dynamics network has been trained, the weighted controllability Gramian from Eq. \ref{eq:linearization} may be ill-conditioned. 
We thus begin with an $\ell_2$ norm and gradually shift to the $G_t^{-1}$ norm.
We also add a small positive term to the diagonal of $G_t$ to ensure invertibility.

\subsection{Learning a Collision Checker}

The third network in our architecture learns to identify if the continuous trajectory between two latent states will be in collision; we denote this network as $\cc(\vec{z}_t, \vec{z}_{t+1}, \xobs)$, parameterized by $\zeta$. 
This is a supervised learning process entailing binary classification; meaning, we assume we have a collision checker in the full state space capable of reporting if the trajectory is in collision -- arguably, a very mild assumption. 
We note here our use of $\xobs$ as an input to the network. In SBMP $\xobs$ is generally only implicitly defined through the collision checker. We use it here to denote any environmental information available to the state space collision checker. This representation can be quite general as the neural network architecture has the capacity to arbitrarily project this information as necessary. For example, this may be workspace coordinates of the obstacles, an occupancy grid, or an image of the problem.
This network is learned separately, after the encoder and dynamics have been learned.
It is possible to learn this network through unsupervised learning by identifying regions of the state space that are rarely or never connected dependent on the environment. We leave this for future work.

\begin{figure*}[htb]
\centering
\includegraphics[width=0.9\textwidth]{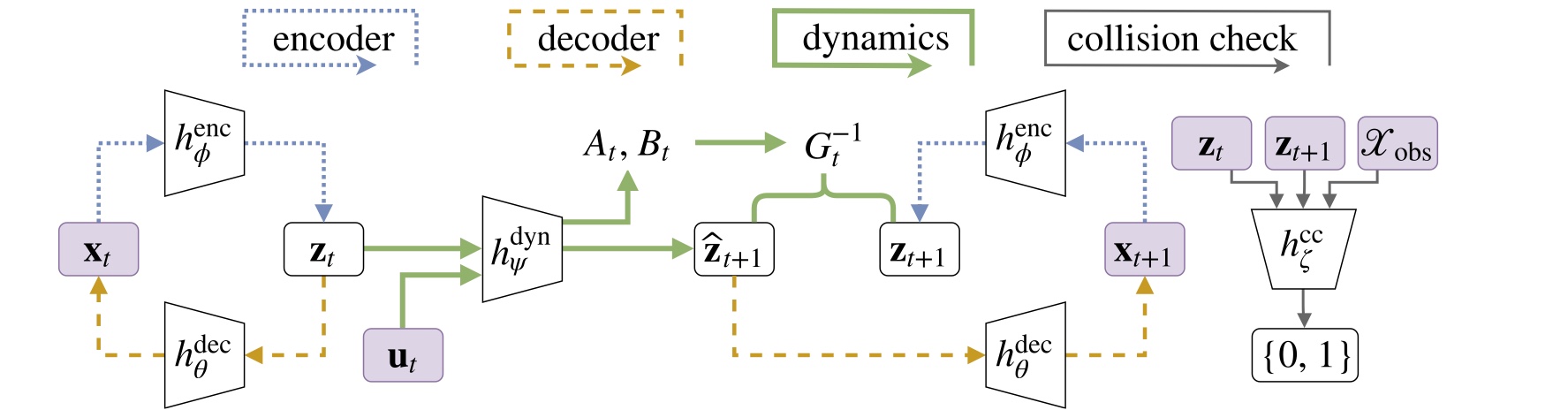}
\caption{The \algname network architecture. The encoder network is shown in blue, the decoder network in orange, the dynamics network in green, and the collision checking network in gray. The inputs are shown in purple. }
\label{fig:arch}
\end{figure*}

\section{Planning in the Latent Space}\label{sec:planning}

We now present the Learned Latent Rapidly-exploring Random Trees (\algtree) algorithm used to globally search the latent space learned as described in Section \ref{sec:latentSpace}.
\algtree is a SBMP algorithm within the rapidly exploring random trees (\rrt) family of algorithms \cite{LaValleKuffner2001}.
Briefly, the algorithm builds a tree, $\mathcal{T}$, through the latent space by sampling latent samples from the learned manifold and propagating dynamics from nearby tree nodes (under the supervision of the collision checking network).
As the learned dynamics may be quite complex, the algorithm must rely on a dynamics propagation strategy, rather than a steering function, to connect local latent states.

\begin{figure}[htb]
    \centering
    \begin{subfigure}{0.15\textwidth}
        \includegraphics[width=\textwidth]{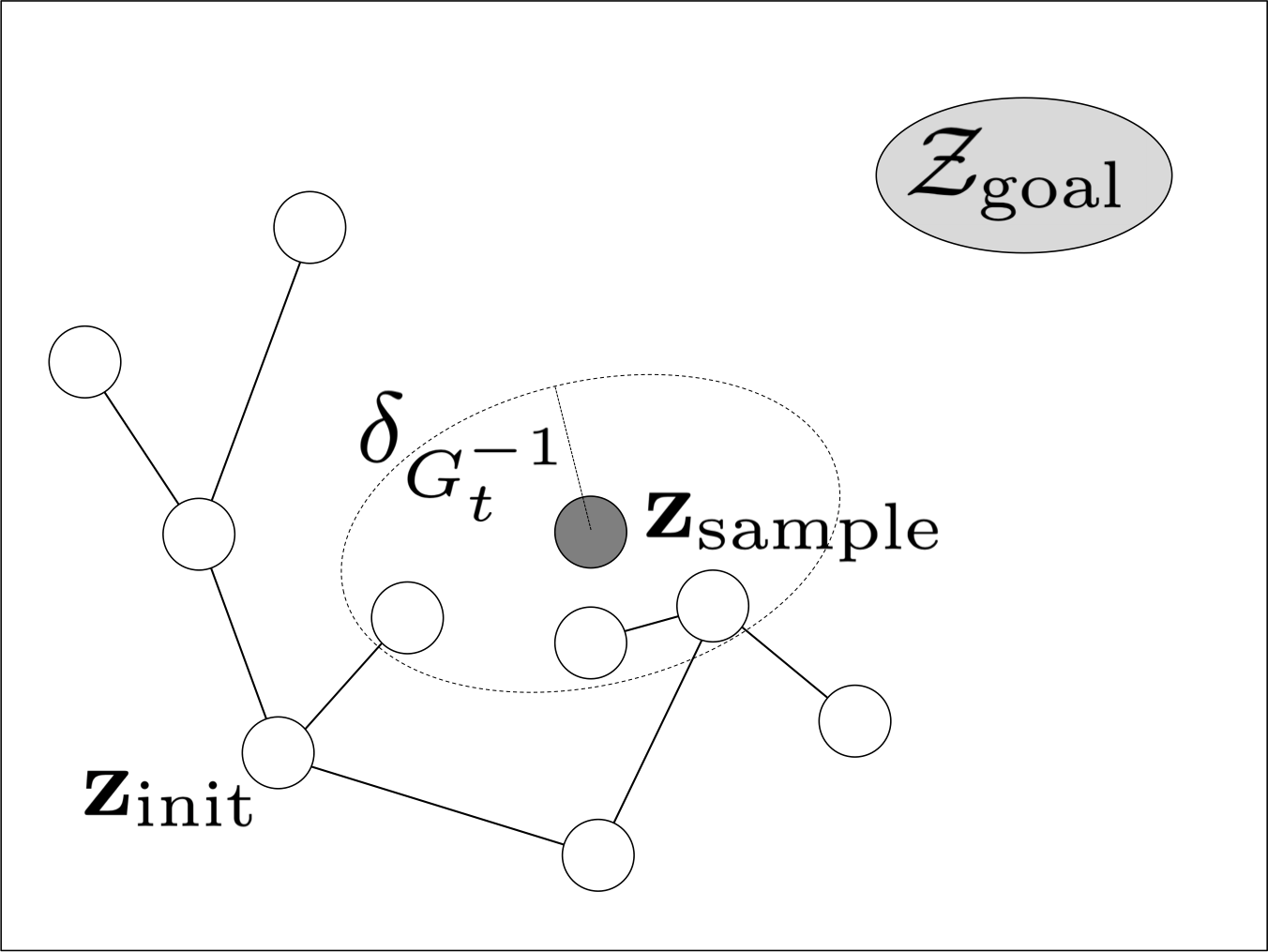}
        \caption{Lines \ref{line:sample}-\ref{line:selectProp}}
        \label{fig:treeSample}
    \end{subfigure}
    \begin{subfigure}{0.15\textwidth}
        \includegraphics[width=\textwidth]{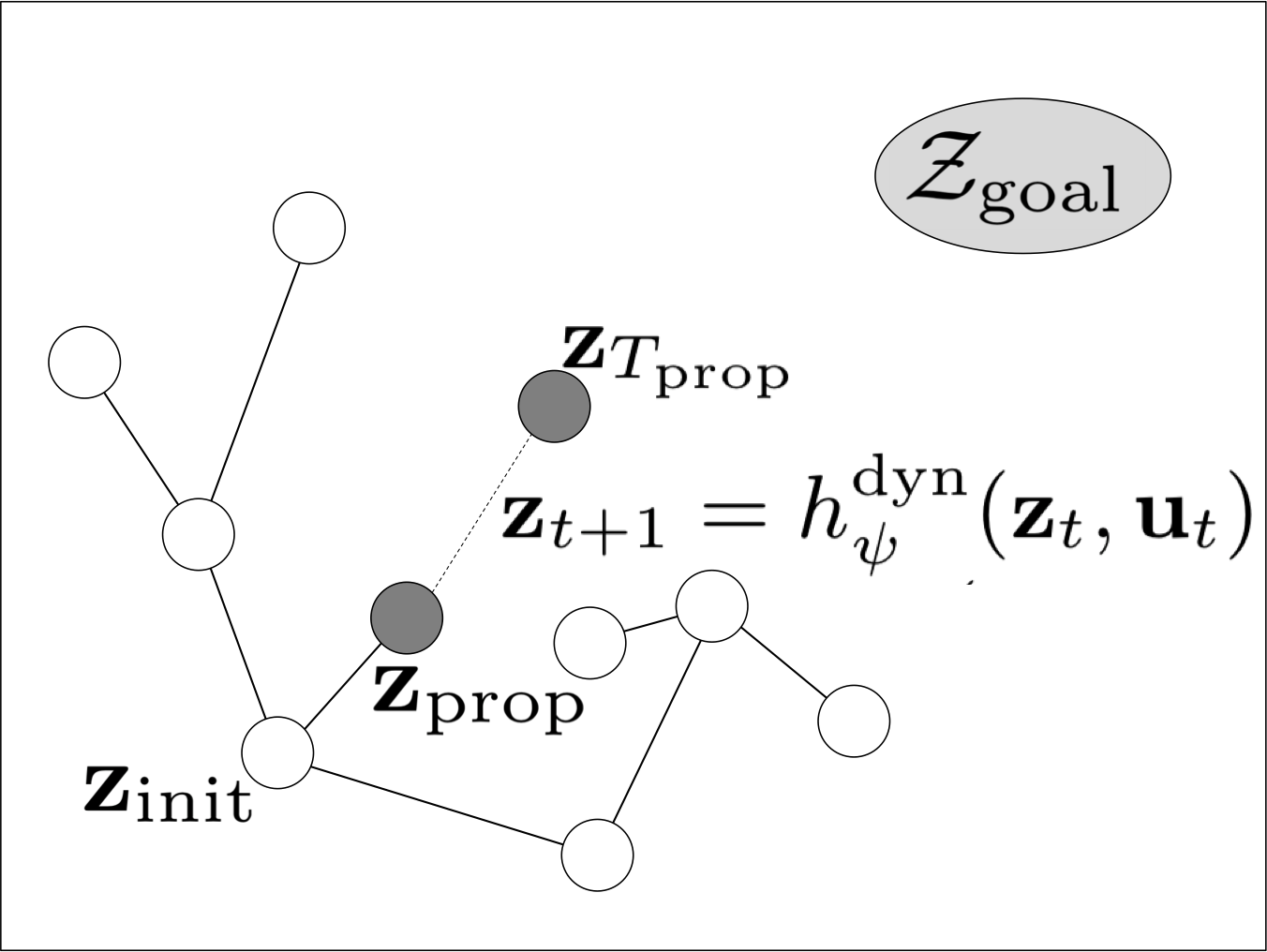}
        \caption{Lines \ref{line:sampleProp}-\ref{line:prop}}
        \label{fig:treeExpand}
    \end{subfigure}
    \begin{subfigure}{0.15\textwidth} 
        \includegraphics[width=\textwidth]{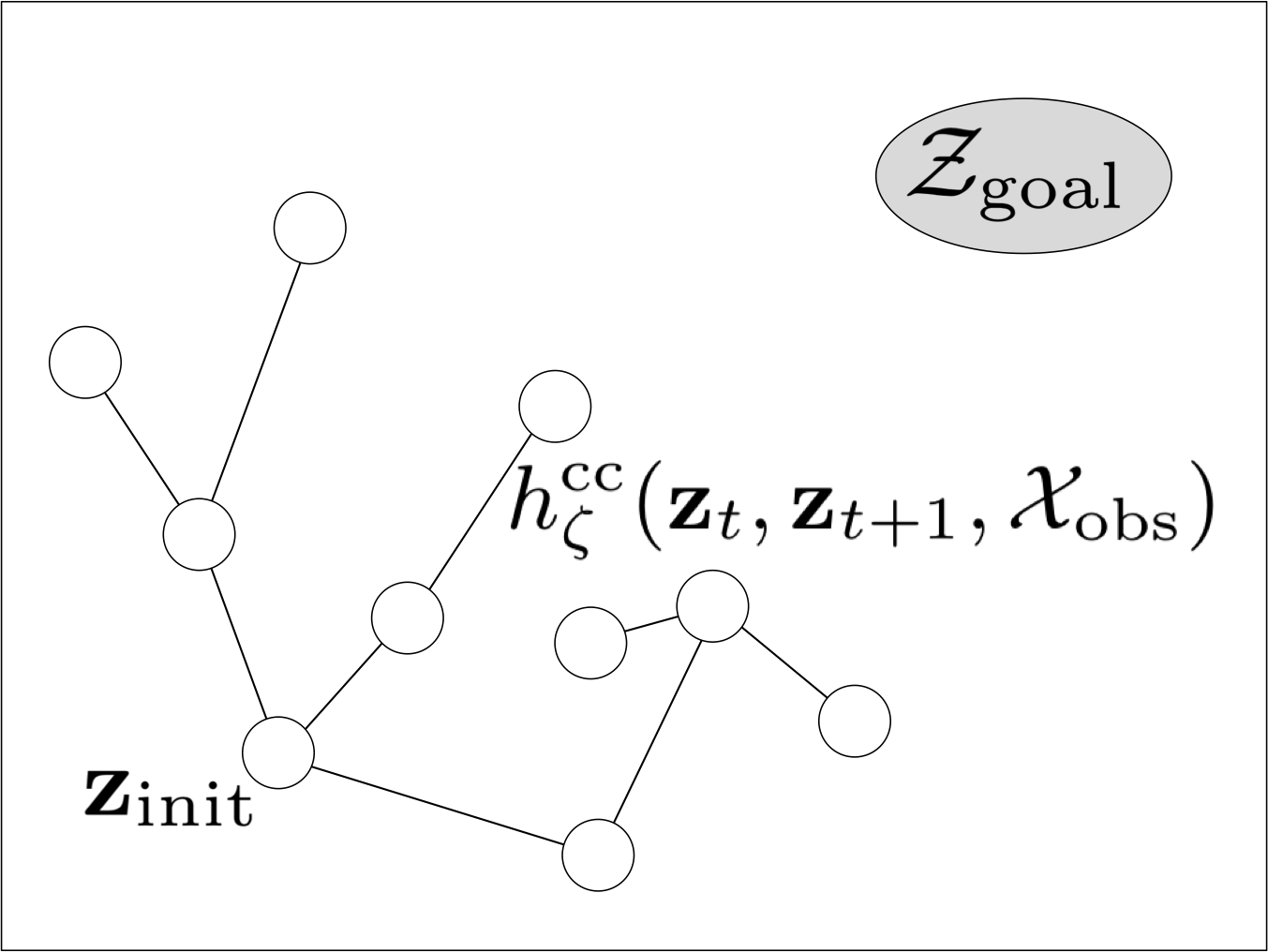}
        \caption{Lines \ref{line:sig}-\ref{line:add}}
        \label{fig:treeCollision}
    \end{subfigure}
   \caption{Learned Latent RRT exploring the latent space, as described in Alg. \ref{alg:treeOutline}. (\ref{fig:treeSample}) Sample $\vec{z}_\text{sample}$ and select lowest-cost tree node within a $\delta_{G_t^{-1}}$ ball. (\ref{fig:treeExpand}) Sample time $T_\text{prop}$ and controls $\vec{u}_{0,\ldots,T_\text{prop}-1}$. Propagate dynamics with $\dyn$. (\ref{fig:treeCollision}) Verify the new edge is collision free with $\cc$. Add the node to the tree.}
\label{fig:treeOutline}
\end{figure}

The \algtree algorithm is similar to the RRT-BestNear algorithm presented in \cite{LiLittlefieldEtAl2016}, with samples restricted to the latent space and dynamic propagation through latent dynamics. Specifically, the algorithm is outlined in Alg. \ref{alg:treeOutline} and visualized in Fig. \ref{fig:treeOutline}.
The algorithm takes as input the planning problem in the full state space: the obstacle representation, $\xobs$ initial state, $\xinit$, and goal region, $\xgoal$.
Note again, we abuse notation here with $\xobs$. Though traditionally in SBMP, $\xobs$ is only implicitly defined through the collision-checker, we use it here to refer to any environmental information used for the state space collision checker.

To begin, the initial state and goal region are encoded into the latent space as $\enc(\xinit) = \zinit$ and $\enc(\xgoal) = \zgoalregion$, respectively (line \ref{line:problem}). 
The tree is initialized with the initial latent state (line \ref{line:initialize}).
The algorithm then begins expanding the tree outwards into the latent space. 
At each iteration a sample, $\vec{z}_\text{sample}$, is selected from a sample set $\z_\text{samples}$ (line \ref{line:sample}).
This set is comprised of latent samples generated by encoding a randomly drawn set of the robot in operation. 
That is, given a sequence of states $(\vec{x}_0,\ldots,\vec{x}_N)$ (where $N$ is large to well cover the operational state space of the robot), we randomly select a sample set $\x_\text{samples}$ and encode it as $\z_\text{samples} = \enc(\x_\text{samples})$.
By selecting samples only from this set, \algtree ensures that the exploration remains nearby the learned manifold in the latent space.
\algtree then selects a node in the tree for propagation by using the \texttt{BestFirstSelection} strategy outlined in \cite{LiLittlefieldEtAl2016} (line \ref{line:selectProp}).
\texttt{BestFirstSelection}$(\vec{z}_\text{sample},\delta_{G_t^{-1}})$ selects the tree node $\vec{z}_\text{prop}$ within a ball radius, $(\vec{z}_\text{prop} - \vec{z}_\text{sample})^\top G_t^{-1}(\vec{z}_\text{prop} - \vec{z}_\text{sample}) < \delta_{G_t^{-1}}$, with the lowest-cost trajectory\footnote{This cost can be fairly general, e.g., control cost or the number of time steps.}. 
If no tree nodes are within the ball radius, the nearest tree node is returned.
The ball radius $\delta_{G_t^{-1}}$ parameter represents a tradeoff between exploration and exploitation of good trajectories; its value is discussed in the subsequent paragraph and at length in \cite{LiLittlefieldEtAl2016}.
With $\vec{z}_\text{prop}$ in hand, \algtree then forward propagates dynamics through $\dyn$ with a randomly chosen control action $\vec{u}_{0,\ldots,T_\text{prop}-1} \in \uspace$ for time $T_\text{prop}$, where $T_\text{prop}$ is randomly chosen between $1$ and $T_{\text{max}}$, following the Monte Carlo propagation procedure outlined in \cite{LiLittlefieldEtAl2016} (line \ref{line:sampleProp}).
This gives a trajectory in latent space $(\vec{z}_0,\vec{u}_0,\ldots,\vec{z}_{T_\text{prop}-1},\vec{u}_{T_\text{prop}-1},\vec{z}_{T_\text{prop}},0)$, where $\vec{z}_0=\vec{z}_\text{prop}$ (line \ref{line:prop}).
Each pair of waypoints is then collision checked by $\cc(\vec{z}_t,\vec{z}_{t+1},\xobs)$ for $t = 0,\ldots,T_\text{prop}-1$.
The probability that a local connection is collision-free can be computed as $\text{sigmoid}(\cc)$ (line \ref{line:sig}). 
Thus, the  acceptable probability threshold $\alpha$ can be tuned to the required level of safety.
Finally, if the propagated trajectory is collision free, $\vec{z}_{T_\text{prop}}$ is added to the tree (line \ref{line:add}).
The exploration ends after a set amount of time or number of samples.
If no trajectories end in the goal region $\zgoalregion$, the algorithm reports failure (line \ref{line:failure}).
If any trajectories reach the goal region, the best trajectory\footnotemark[1] is then selected and projected back to the full state space through $\dec$ (line \ref{line:selectTraj}-\ref{line:decode}).
This outputs nominal control inputs as well as the full state space trajectory (which may be used as input to a controller or a trajectory optimizer).

\begin{algorithm}
\caption{Learned Latent RRT Outline}
\label{alg:treeOutline}
\algsetup{linenodelimiter=}
\begin{algorithmic}[1]
\ENSURE $\delta_{G_t^{-1}}$, $T_{\max}$, $\z_\text{samples}$, $\alpha$
\REQUIRE ($\xinit$, $\xgoal$, $\xfree$)
\STATE Encode problem input into latent space: \\ $\zinit = \enc(\xinit)$, $\zgoalregion = \enc(\xgoal)$ \label{line:problem}
\STATE Add $\zinit$ to the tree $\mathcal{T}$ \label{line:initialize}
\WHILE{time remains}
\STATE Sample $\vec{z}_\text{sample} \in \z_\text{samples}$ \label{line:sample}
\STATE $\vec{z}_\text{prop} = \texttt{BestFirstSelection}(\vec{z}_\text{sample},\delta_{G_t^{-1}})$ \label{line:selectProp}
\STATE Sample $T_\text{prop} \in \{1,\ldots,T_{\max}\}$ and $\vec{u}_{0,\ldots,T_\text{prop}-1} \in \uspace$ \label{line:sampleProp}
\STATE Propagate dynamics $\vec{z}_{t+1} = \dyn(\vec{z}_t,\vec{u}_t)$ from $\vec{z}_0=\vec{z}_\text{prop}$ for $T_\text{prop}$ time steps, to generate latent trajectory $(\vec{z}_0,\vec{u}_0,\ldots,\vec{z}_{T_\text{prop}-1},\vec{u}_{T_\text{prop}-1},\vec{z}_{T_\text{prop}},0)$ \label{line:prop}
\IF{$\text{sigmoid}(\cc(\vec{z}_t,\vec{z}_{t+1},\xobs)) > \alpha \, \forall \, t$}\label{line:sig}
\STATE Add $\vec{z}_T$ to the tree $\mathcal{T}$ \label{line:add}
\ENDIF
\ENDWHILE
\IF{$\mathcal{T} \cap \zgoalregion = \emptyset$}
\STATE Report failure \label{line:failure}
\ELSE 
\STATE Select lowest-cost trajectory with its end in $\mathcal{T} \cap \zgoalregion$\label{line:selectTraj}
\STATE Project back to full state space through $\dec$ and return $(\xinit,\vec{u}_0,\ldots,\vec{x}_T,\vec{u}_T)$, where $T$ is time steps to $\zgoal$\label{line:decode}
\ENDIF 

\end{algorithmic}
\end{algorithm}

Briefly we provide discussion on how to choose values for the algorithm's parameters.
The $\delta_{G_t^{-1}}$ ball radius represents a tradeoff between exploiting good trajectories and exploring with new ones.  
This value should be tuned according to the system and from discussion in \cite{LiLittlefieldEtAl2016}.
The maximum dynamic propagation time $T_{\max}$ controls, roughly, the length of edge.
For subsequent results this was approximately set so the robot with optimal control can traverse the full state space in $4T_{\max}$.
The value of $\alpha$ reflects the required level of safety in the collision checking network.
We found most collision classifications yielded probabilities less than 1\% or greater than 99\%, however, for subsequent results this threshold was set as $\alpha = 0.9$.
Finally, we also use a technique called goal-biasing, in which we sample more heavily from the goal region. 
For subsequent results we sample a state within the goal region 10\% of the time. 

Due to the approximate nature of the learned latent space, \algtree does not enjoy formal completeness or optimality guarantees. An interesting avenue for future research is to investigate conditions under which theoretical guarantees for \algtree can be derived. 

\section{Experimental Results}\label{sec:experiments}

In this section we demonstrate the effectiveness of our methodology on two planning problems from domains in which traditional motion planning approaches are intractable.
The first is a visual planning problem entailing a point robot with single integrator dynamics navigating a cluttered environment based on visual inputs. The second is a humanoid planning problem where a humanoid robot has to maneuver amidst obstacles. The methodology was implemented in Python and Tensorflow \cite{Abadi2015}. All data was collected on a Unix system with a 3.4 GHz CPU and an NVIDIA GeForce GTX 1080 Ti GPU.
Example code can be found at \url{github.com/StanfordASL/LSBMP}.

\subsection{Visual Planning Problem}

We consider a visual space planar problem similar to the one outlined in \cite{watter2015embed}, though with a more complex obstacle space. This setup allows us to demonstrate the performance of the \algname methodology to generalize to new environments by leveraging the algorithmic prior of separating collision checking from the dynamic propagation process. 
As we have access to the true underlying state, and the underlying planning problem is relatively simple, this problem allows us to compare our methodology against sampling-based motion planning algorithms with access to the true state of the system.

\begin{figure}[htb]
    \centering
    \begin{subfigure}{0.20\textwidth}
        \includegraphics[width=\textwidth]{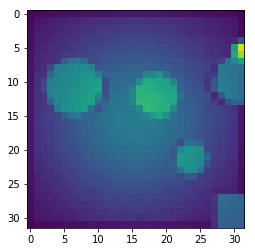}
        \caption{Initial state}
        \label{fig:init355}
    \end{subfigure}
    \begin{subfigure}{0.20\textwidth}
        \includegraphics[width=\textwidth]{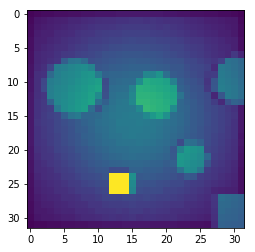}
        \caption{Goal state}
        \label{fig:goal355}
    \end{subfigure}
    \begin{subfigure}{0.24\textwidth} 
        \includegraphics[width=\textwidth]{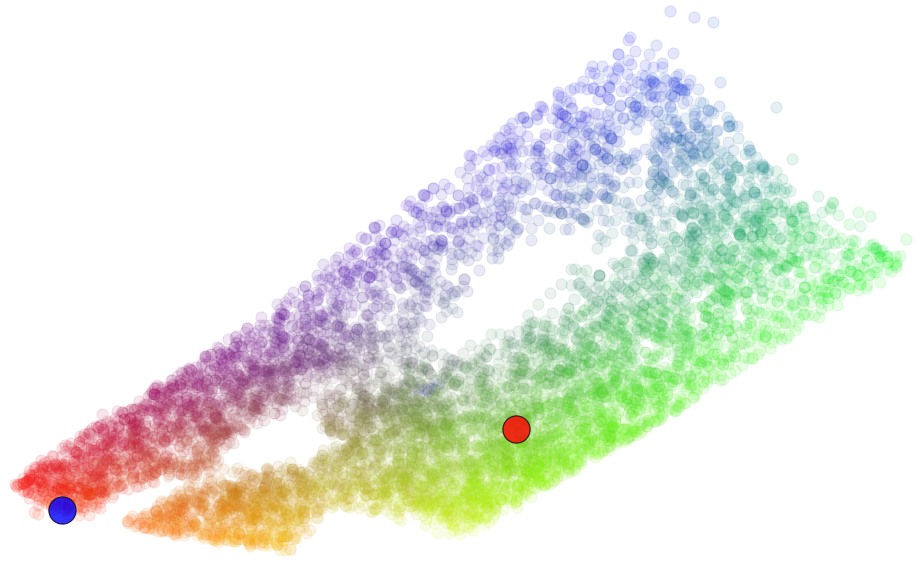}
        \caption{Latent space}
        \label{fig:geometricSpace}
    \end{subfigure}
    \begin{subfigure}{0.21\textwidth} 
        \includegraphics[width=\textwidth]{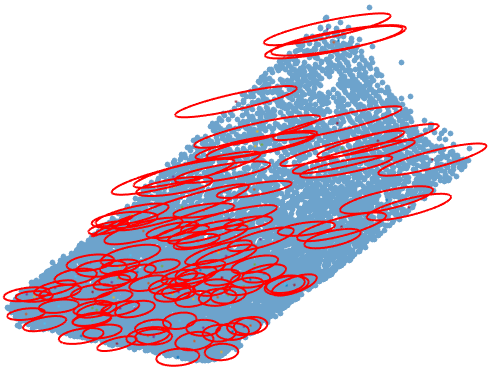}
        \caption{$G_t^{-1}$}
        \label{fig:geometricGramian}
    \end{subfigure}
   \caption{(\ref{fig:init355}-\ref{fig:goal355}) The full state of the visual robot with obstacles (robot is shown in yellow) and initial and goal states for input to our methodology. (\ref{fig:geometricSpace}) The learned latent space colored by the position of the robot in the true underlying state space. States from the lower-left corner are green, lower-right are yellow, upper-right are red, and upper-left are blue. The encoded initial state is shown in blue and the goal state in red. States in collision are not visualized (empty white regions). (\ref{fig:geometricGramian}) The inverse of the weighted controllability Gramian in the state space (an ellipse representing equal energy costs). The learned latent space in the upper right is less dense so the ellipses are larger in the direction of the spread, thus penalizing errors less.}
\label{fig:geometricProblem}
\end{figure}

The full state of the visual problem is a $32\times32$ pixel image rendered in MuJoCo showing both the robot and the obstacles in the scene.
The robot is a point robot with single integrator dynamics.
The obstacles are a random number of circles and squares with randomly generated locations and sizes.
An example problem's initial and goal states are shown in Figs. \ref{fig:init355}-\ref{fig:goal355}.
The training data for the autoencoder and dynamics network consists of 10,000 environments, each with a trajectory of 10 successive states and control inputs.
The training data for the collision checking network consists of 25,000 environments, again with 10 pairs of states and a collision label. 

The full state is projected down to a two dimensional latent space.
The methodology applied to an image-based system requires some architectural changes to make the learning problem more tractable.
The encoder network, $\enc$, is a deep-spatial autoencoder \cite{finn2016deep}, which encourages learning important visual features by using a convolutional neural network followed by a spatial soft arg-max.
To create accurate reconstructions of the full state the decoder network is given as input an image of the environment in addition to the latent state.
This additional input image may be the environment without the robot present or it could be an image with the robot from a different time step (e.g., the initial state for a planning problem).

Fig. \ref{fig:geometricSpace} shows the learned latent space for the problem in Fig. \ref{fig:init355}-\ref{fig:goal355}. 
The methodology learns a latent space that captures the position of the robot and effectively removes latent regions within obstacles, though this obstacle set has never been seen before. 
Fig. \ref{fig:geometricGramian} shows the inverse weighted controllability Gramian in the latent space. 
Because the true system dynamics are consistent throughout the state space, this gives an opportunity to evaluate the weighted loss function.
As desired, in regions of the state space that are more spread (i.e., the same control input allows more movement), errors are penalized less.

\begin{figure}[htb]
    \centering
    \begin{subfigure}{0.20\textwidth}
        \includegraphics[width=\textwidth]{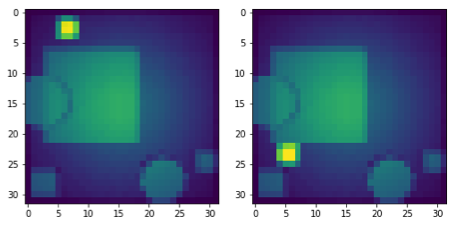}
        \caption{Initial and goal states}
        \label{fig:initGoal43}
    \end{subfigure}
    \begin{subfigure}{0.20\textwidth}
        \includegraphics[width=\textwidth]{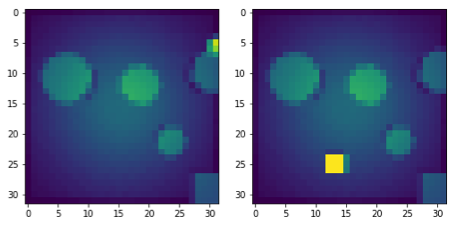}
        \caption{Initial and goal states}
        \label{fig:initGoal355}
    \end{subfigure}
    \begin{subfigure}{0.21\textwidth}
        \includegraphics[width=\textwidth]{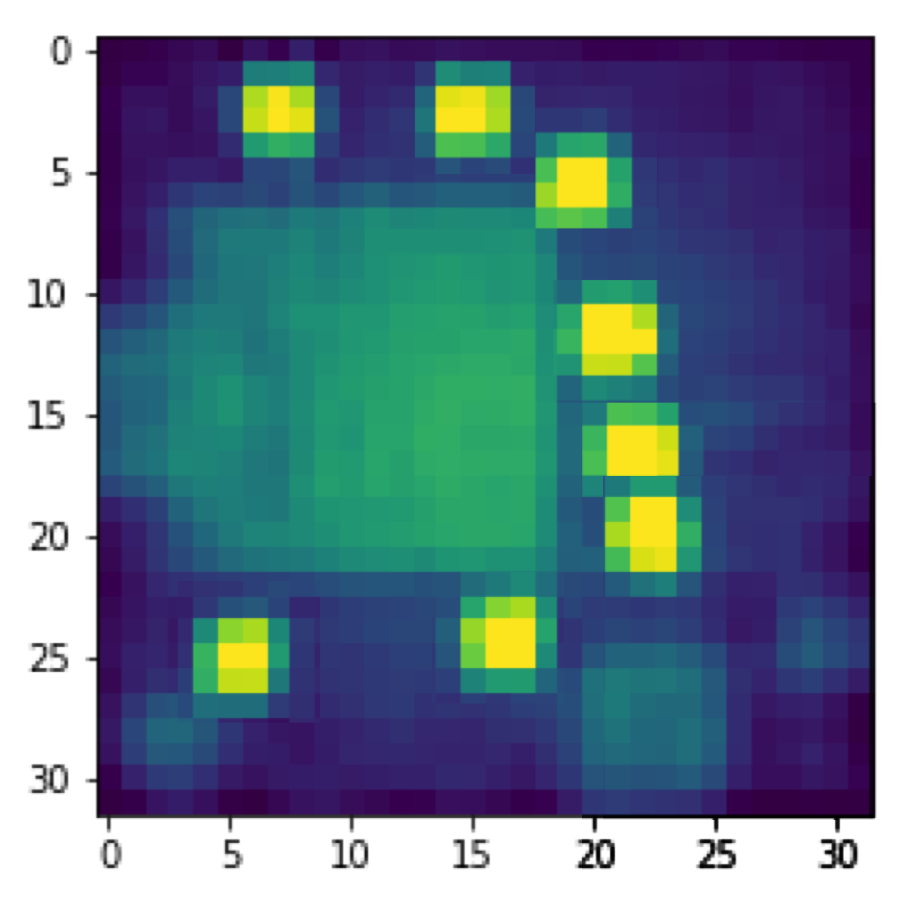}
        \caption{Decoded trajectory}
        \label{fig:path43}
    \end{subfigure}
    \begin{subfigure}{0.21\textwidth}
        \includegraphics[width=\textwidth]{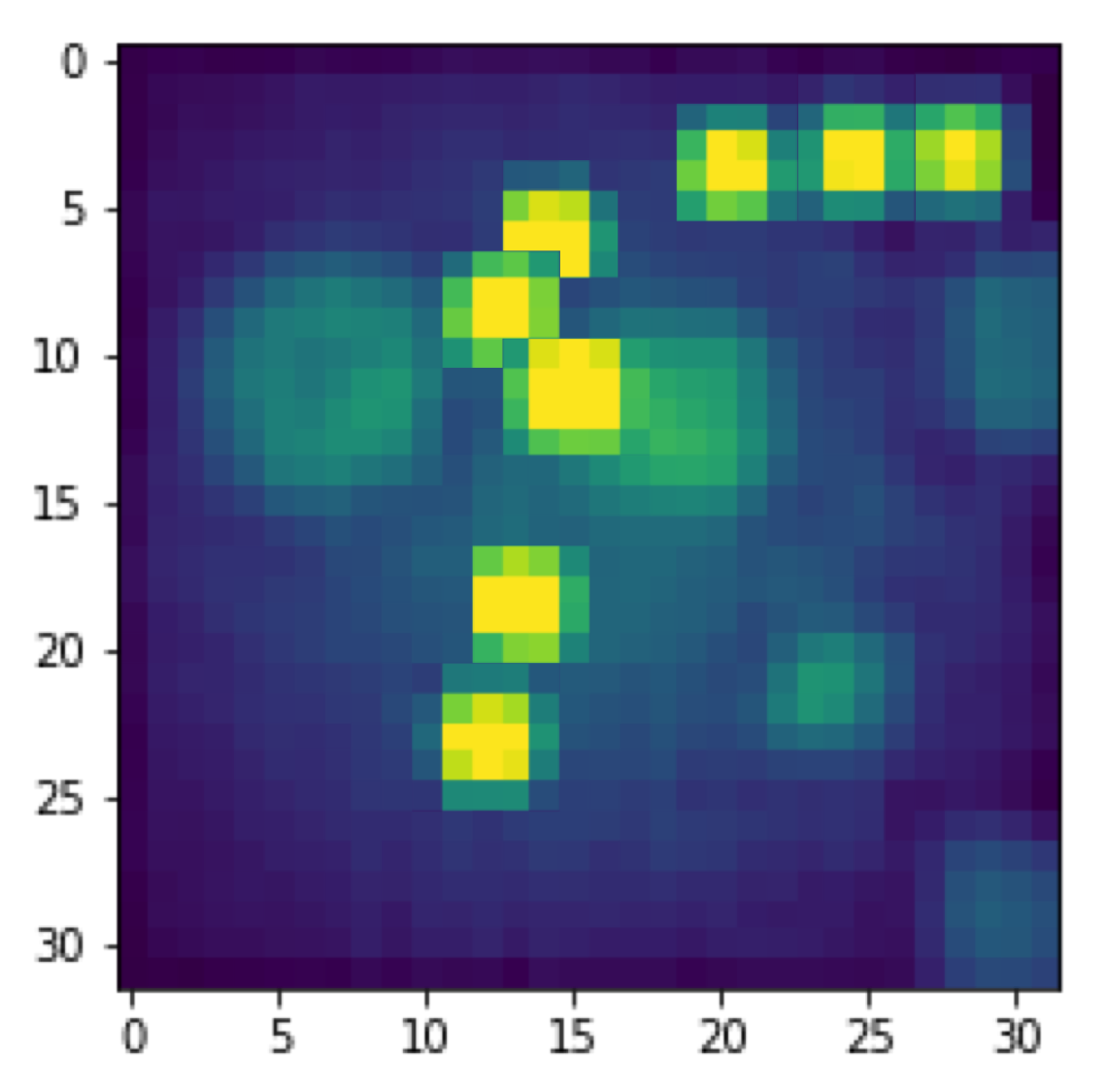}
        \caption{Decoded trajectory}
        \label{fig:path355}
    \end{subfigure}
    \begin{subfigure}{0.23\textwidth}
        \includegraphics[width=\textwidth]{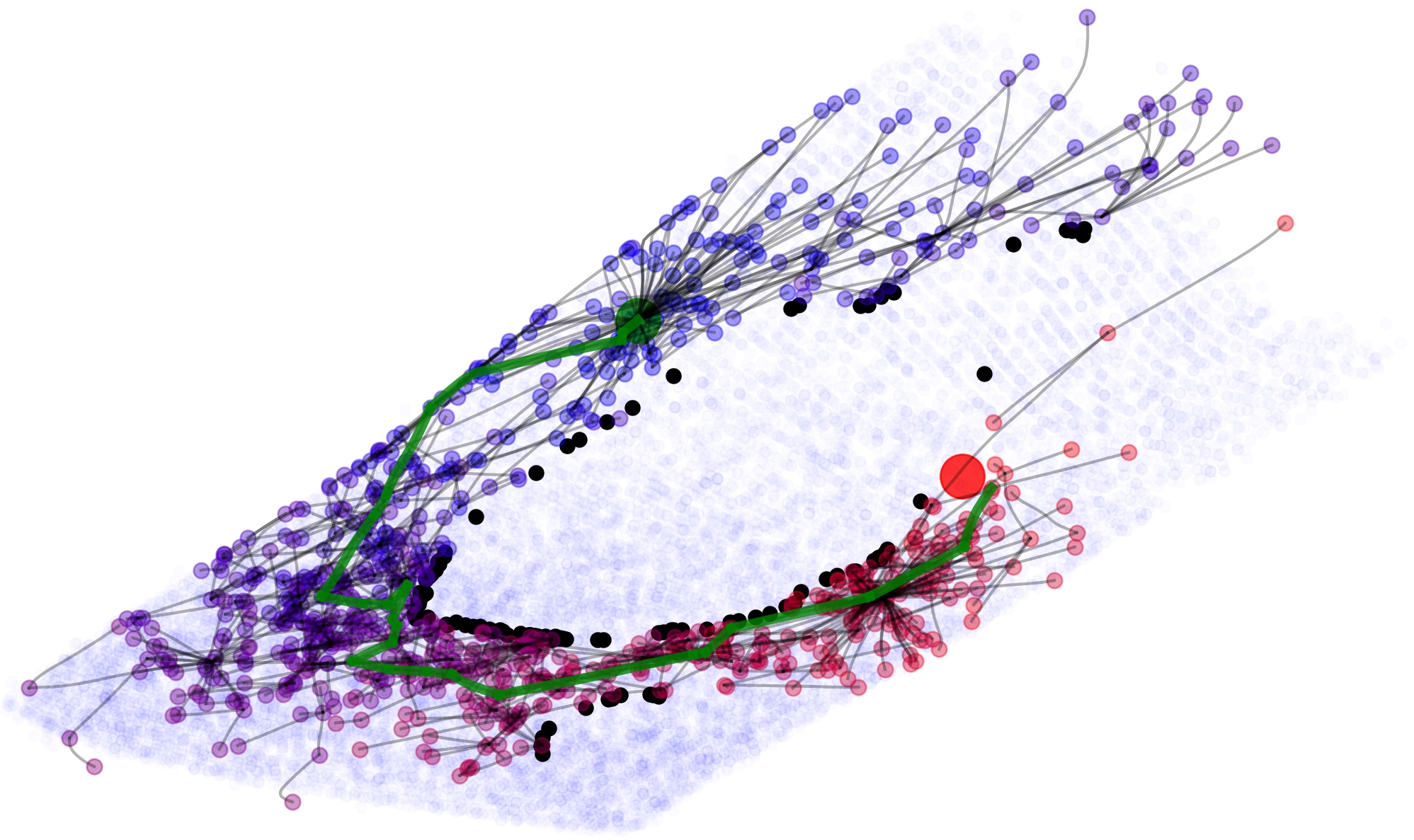}
        \caption{Latent space \algtree}
        \label{fig:tree43}
    \end{subfigure}
    \begin{subfigure}{0.23\textwidth}
        \includegraphics[width=\textwidth]{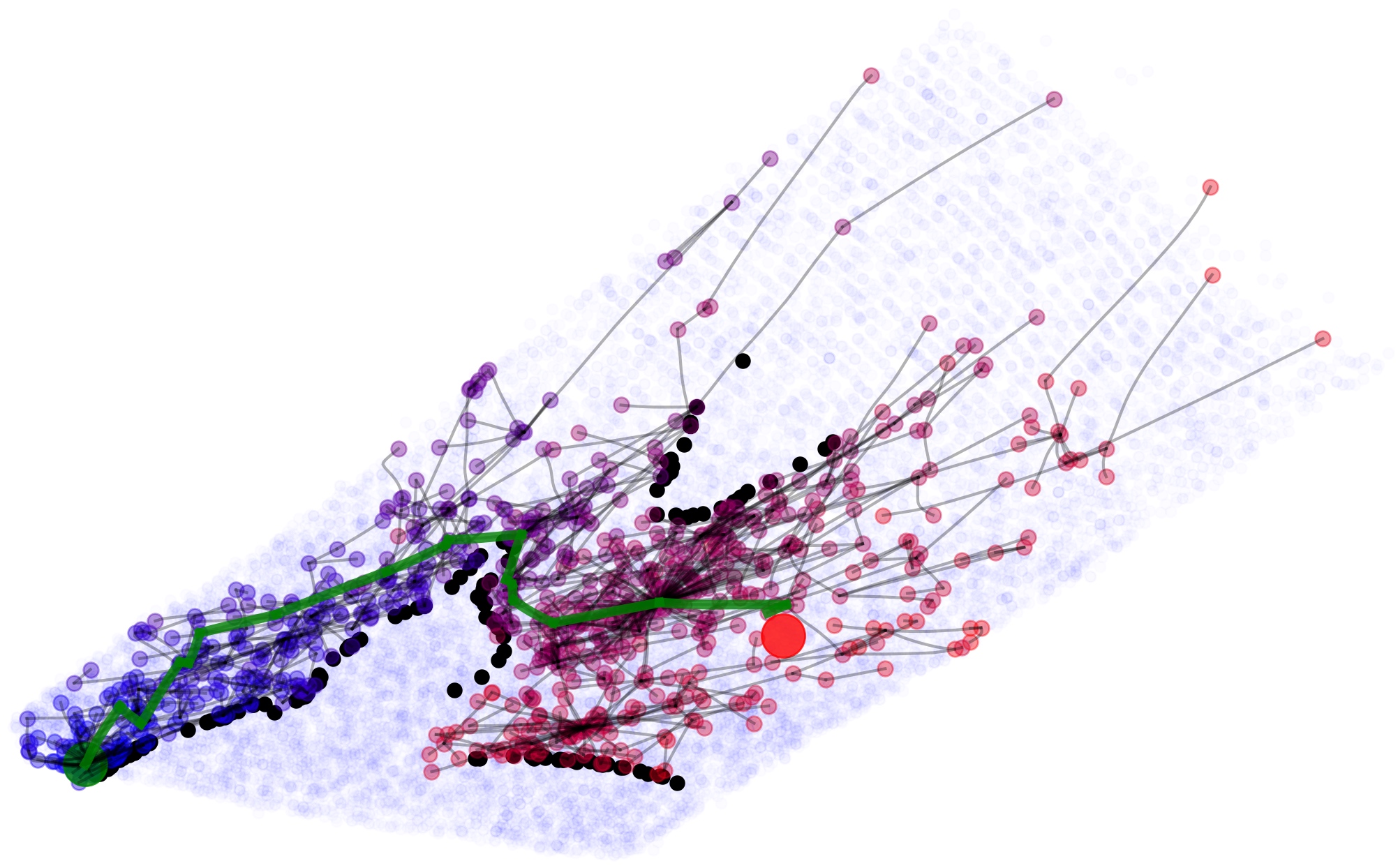}
        \caption{Latent space \algtree}
        \label{fig:tree355}
    \end{subfigure}
   \caption{(\ref{fig:initGoal43}-\ref{fig:initGoal355}) Initial and goal states for the visual planning problem. (\ref{fig:path43}-\ref{fig:path355}) Planned trajectories decoded into the full state space. Note that collisions for this system are only determined by the center of the point robot. (\ref{fig:tree43}-\ref{fig:tree355}) \algtree in the latent space, where the initial state and final trajectory are shown in green, the goal state in red, and collisions between states in black. States within the tree are color-coded by their cost (blue denotes low cost, red denotes high cost). Lastly, samples from the training data are used as samples in \algtree and are shown in blue. }
\label{fig:geometricFull}
\end{figure}

Fig. \ref{fig:geometricFull} shows two planning problems for this system, the planned trajectories in the full state space, and the \algtree in latent space. 
\algname is able to successfully generalize to these problems and find trajectories even through narrow passages.
These global plans can then be fed into a local controller, such as \cite{banijamali2017robust}, or a trajectory optimizer \cite{srinivas2018universal}, to be executed.

\begin{figure}[htb]
    \centering
    \begin{subfigure}{0.22\textwidth}
        \includegraphics[width=\textwidth]{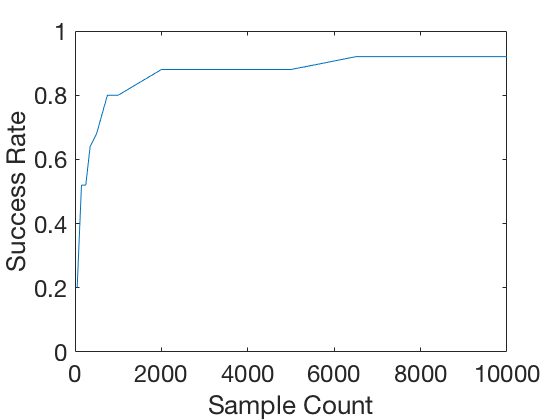}
        \caption{Success rate}
        \label{fig:geometricSuccess}
    \end{subfigure}
    \begin{subfigure}{0.22\textwidth}
        \includegraphics[width=\textwidth]{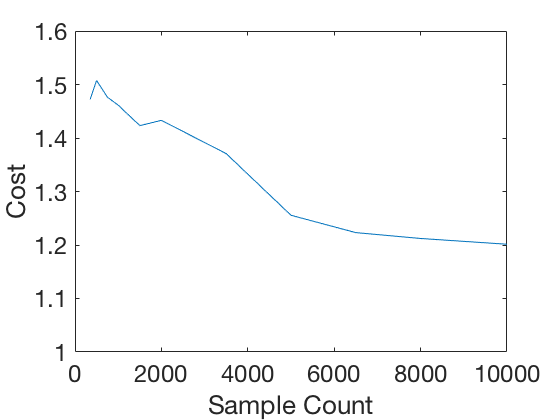}
        \caption{Cost}
        \label{fig:geometricCost}
    \end{subfigure}
    \begin{subfigure}{0.22\textwidth}
        \includegraphics[width=\textwidth]{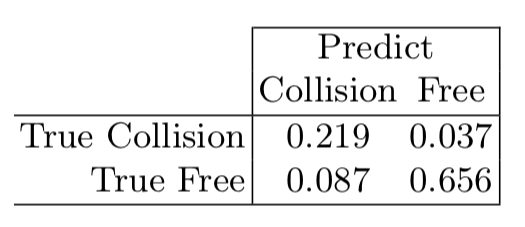}
        \caption{Collision predictions}
        \label{fig:geometricConfusion}
    \end{subfigure}
   \caption{Convergence curves and collision confusion table for visual planning problem.}
\label{fig:geometricConvergence}
\end{figure}

Fig. \ref{fig:geometricConvergence} shows the success rate and cost convergence curve (in terms of control input, i.e., distance) versus the number of samples. The success rate increases quickly at first before leveling below 100\% (though all problems are solvable). The cost convergence curve follows a standard curve for SBMP, converging quickly at first before leveling out. The confusion matrix shows nearly 90\% of collisions are correctly classified, with only 4\% of true collisions classified as free. This false positive ratio could likely be reduced by including conservatism in the full state space collision checker.
With 1000 samples, the algorithm takes on average 9 seconds.

We additionally compare these plans to plans computed in the true state space (given a perfect representation of the robot and obstacles in two dimensions) by the Fast Marching Tree (\fmt) algorithm \cite{JansonSchmerlingEtAl2015} and by RRT-BestNear \cite{LiLittlefieldEtAl2016} (which uses the same exploration strategy as \algtree, but with knowledge of the true state space). 
We note that \fmt relies on a two point boundary value problem solver to make local connections.
Each algorithm is run with 2000 samples and 100 randomly generated planning problems.
This allows \fmt to explore the state space more efficiently than \algtree or RRT-BestNear (which only use dynamics propagation).
In terms of success rate of finding a solution, normalized to \fmt, \algtree is able to solve 92\% of the problems while RRT-BestNear solves 96\%.
In terms of solution trajectory cost, \algtree solves these problems with a 13\% higher cost than \fmt and RRT-BestNear solves with a 5\% higher cost than \fmt.
From these results we make two observations.
First, planning in the latent space incurs some loss (in terms of success rate and cost), however, this loss is relatively small considering the algorithm is only given a very high-dimensional, inexact, visual representation of the state.
Second, if the latent space were forced into a more restrictive class of dynamical systems (e.g., locally-linear) with a two point boundary value problem solver available, the \algname methodology could leverage algorithms like \fmt to more efficiently explore the state space.
We leave this to future work.

\subsection{Humanoid Robot}

Our second numerical experiment shows the methodology with a high-dimensional dynamical system, a humanoid in OpenAI Roboschool \cite{brockman2016openai}.
The system consists of 50 states with dynamics and a learned controller from OpenAI Roboschool, which takes as input a target position.
The obstacles are randomly generated spheres or cubes of random sizes and positions.
The initial and goal states too are randomly generated through the state space.

The state is encoded into a four dimensional latent space, which was chosen inspired by \cite{ha2017high}.
The network architecture is fully connected.
The collision checking network takes as input two latent states and a vector of the centers of each obstacle, the type of obstacle (sphere or cube), and the sphere radius or cube width of the obstacle. 
The autoencoder and dynamics network was trained on 1,000,000 data points (states and control input), sampled at 10~Hz. 
This data was obtained by randomly sampling control inputs (target positions) and propagating dynamics with the Roboschool controller \cite{brockman2016openai}, essentially randomly exploring the operational state space.
The collision checking network was trained on 1,000,000 data points (pairs of states and the location, type, and size of each obstacle in the workspace).

\begin{figure}[htb]
    \centering
    \begin{subfigure}{0.1115\textwidth}
        \includegraphics[width=\textwidth]{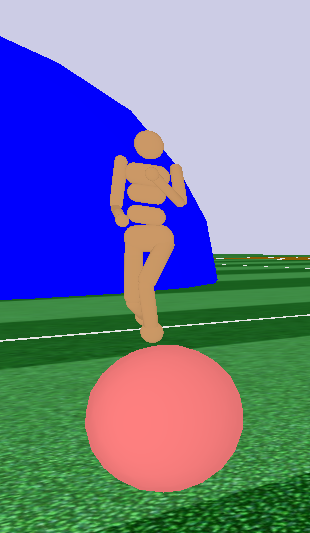}
        \caption{}
         \label{fig:humanoid}
    \end{subfigure}
    \begin{subfigure}{0.366\textwidth}
        \includegraphics[width=\textwidth]{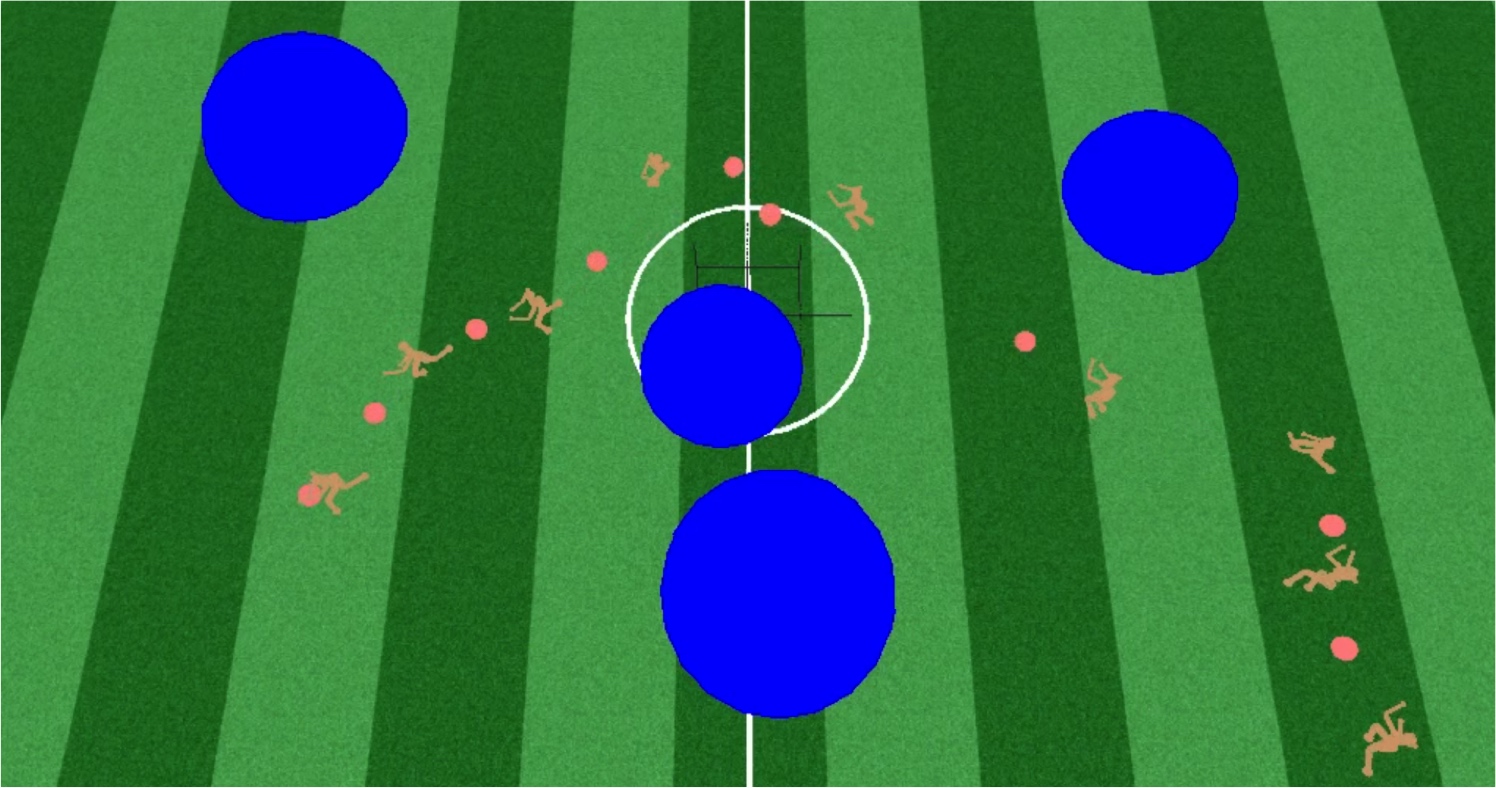}
        \caption{}
        \label{fig:humanoidRun}
    \end{subfigure}
   \caption{(\ref{fig:humanoid}) The humanoid robot with an obstacle in blue and the control target in red. (\ref{fig:humanoidRun}) The humanoid robot executing the planned trajectory in the full state space (video at \texttt{goo.gl/wrm9JR}).}
\label{fig:humanoidReal}
\end{figure}

\begin{figure}[htb]
    \centering
    \begin{subfigure}{0.15\textwidth}
        \includegraphics[width=\textwidth]{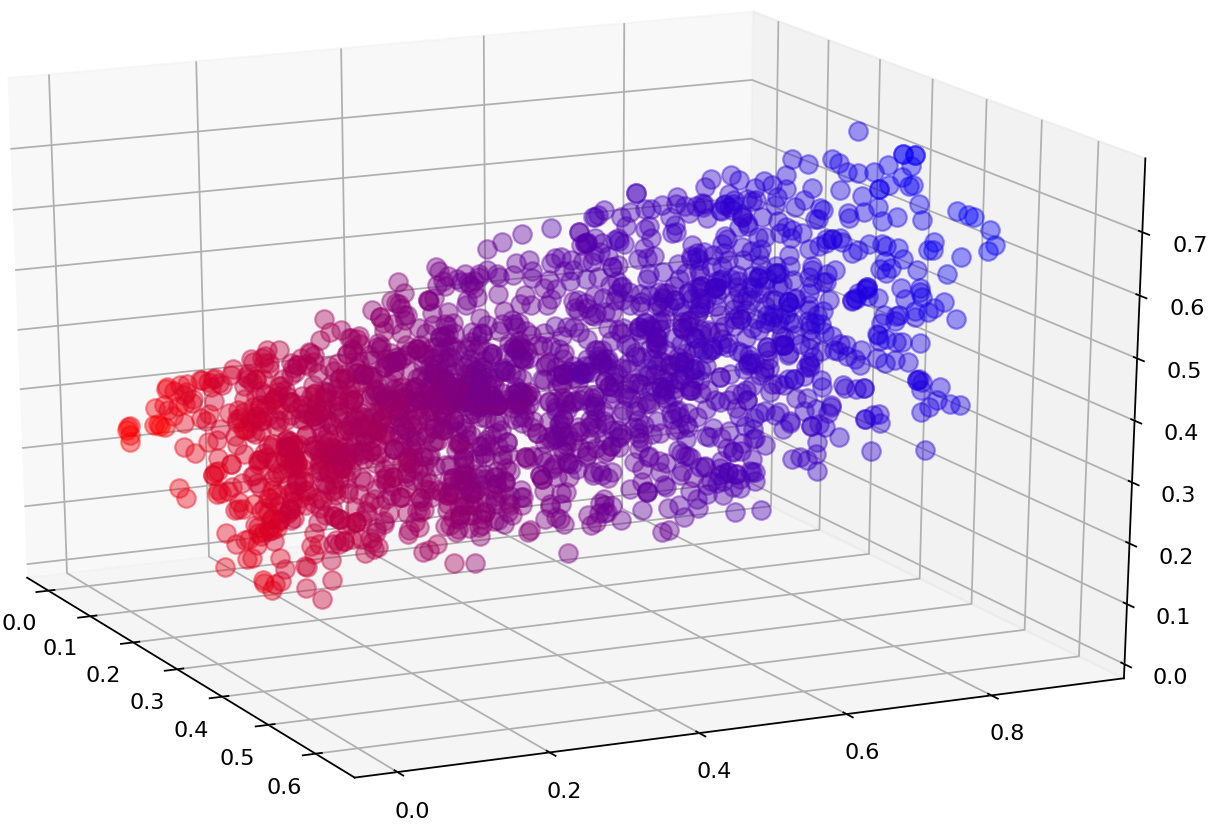}
        \caption{$x$ center of mass}
        \label{fig:humanoidX}
    \end{subfigure}
    \begin{subfigure}{0.15\textwidth}
        \includegraphics[width=\textwidth]{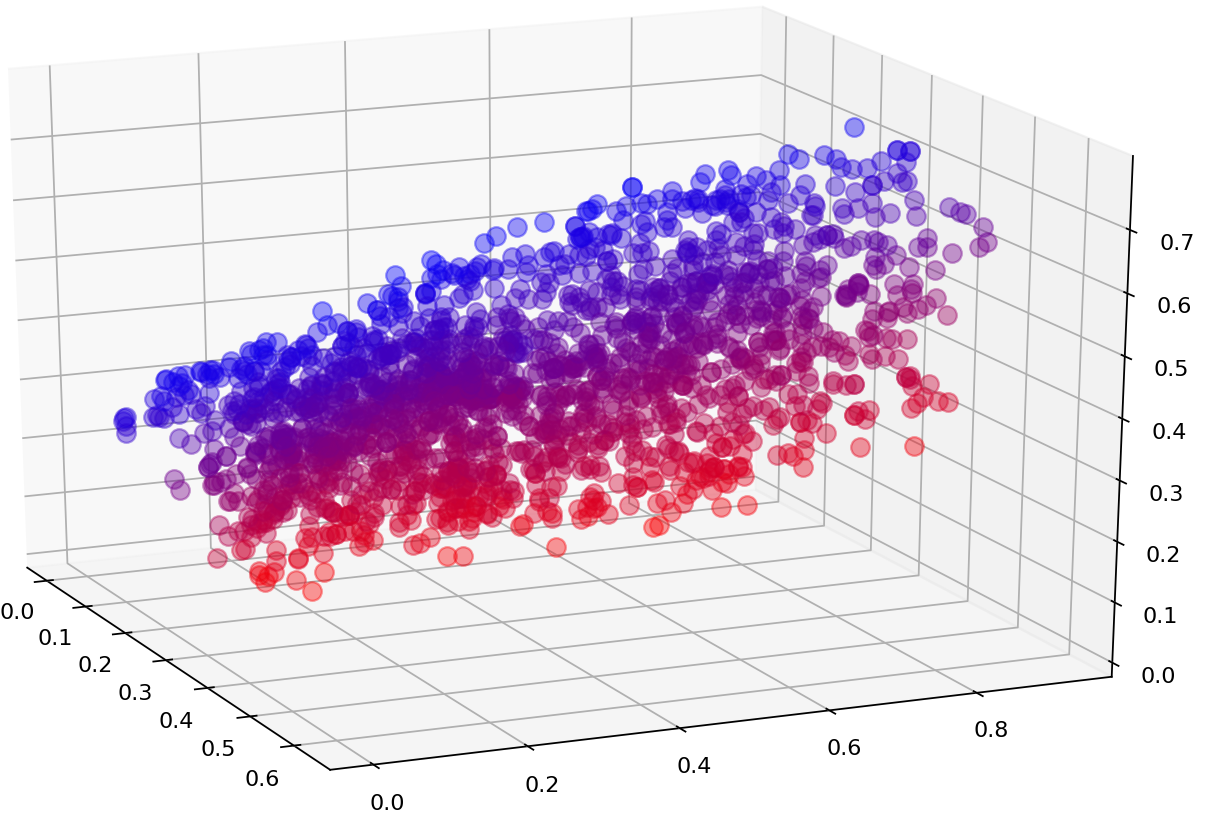}
        \caption{$y$ center of mass}
        \label{fig:humanoidY}
    \end{subfigure}
    \begin{subfigure}{0.15\textwidth}
        \includegraphics[width=\textwidth]{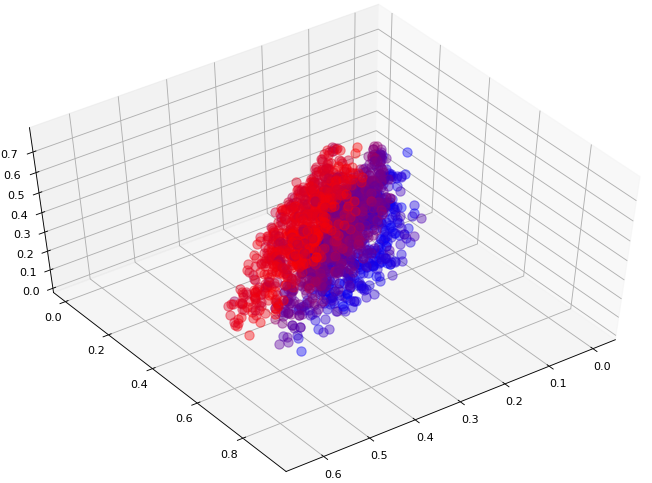}
        \caption{yaw}
        \label{fig:humanoidYaw}
    \end{subfigure}
   \caption{A 3D representation of the structure learned within the humanoid latent space, with each plot visualized by the robot's $x,y,\text{yaw}$ respectively (where red denotes a low value and blue a high value).}
\label{fig:humanoidLatentSpace}
\end{figure}

Fig. \ref{fig:humanoidLatentSpace} shows the structure learned within the latent space.
The latent space captures the intuitively important dynamics of the problem: the robot's position and yaw.
These states most significantly govern the humanoid's dynamics, i.e., it is often traveling in a straight line towards the control target.
The other dimensions of the state space, such as joint angles and velocities, are fairly structured within the gait of the humanoid and only vary small amounts.

\begin{figure}[htb]
    \centering
    \begin{subfigure}{0.245\textwidth}
        \includegraphics[width=\textwidth]{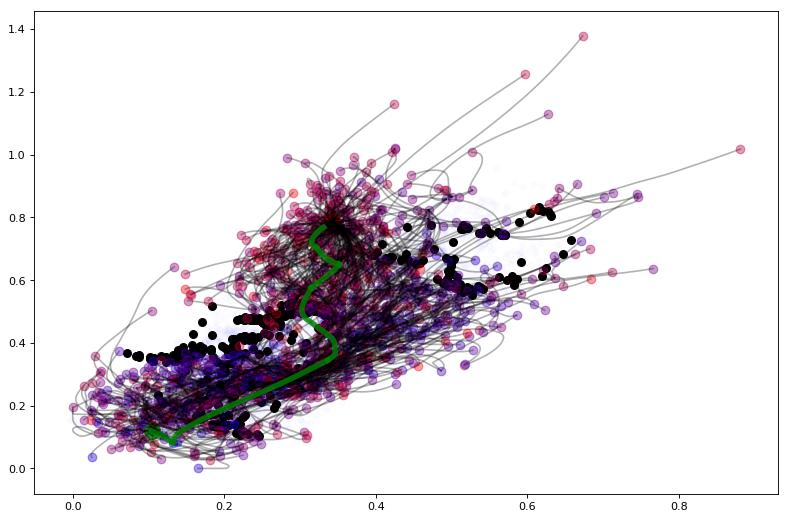}
        \caption{\algtree in 2D latent space}
         \label{fig:humanoidRRT2d}
    \end{subfigure}
    \begin{subfigure}{0.225\textwidth}
        \includegraphics[width=\textwidth]{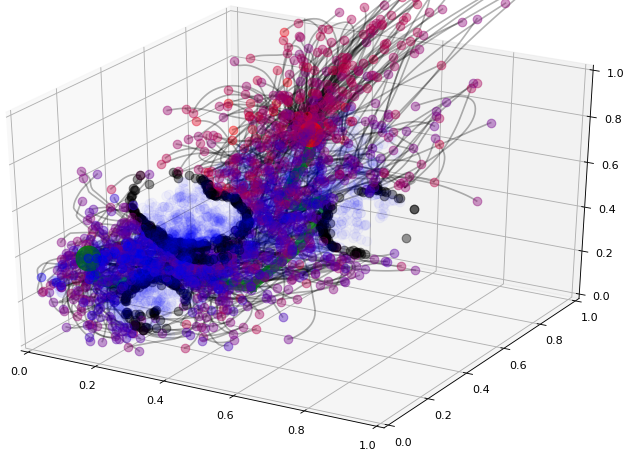}
        \caption{\algtree in 3D latent space}
        \label{fig:humanoidRRT3d}
    \end{subfigure}
   \caption{\algtree in latent space with nodes colored by cost (blue is low cost, red is high cost), collisions are shown in black, and the final trajectory is shown in green.}
\label{fig:humanoidRRT}
\end{figure}

\begin{figure}[htb]
    \centering
    \begin{subfigure}{0.22\textwidth}
        \includegraphics[width=\textwidth]{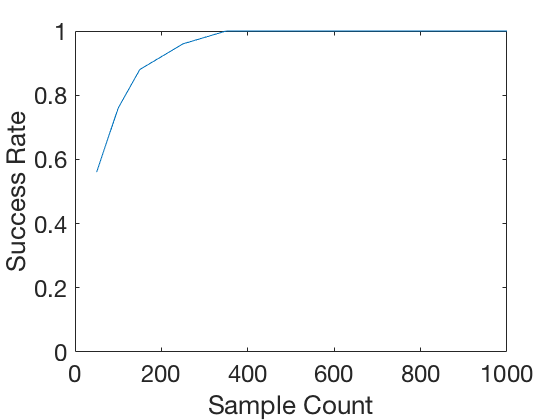}
        \caption{Success rate}
        \label{fig:humanoidSuccess}
    \end{subfigure}
    \begin{subfigure}{0.22\textwidth}
        \includegraphics[width=\textwidth]{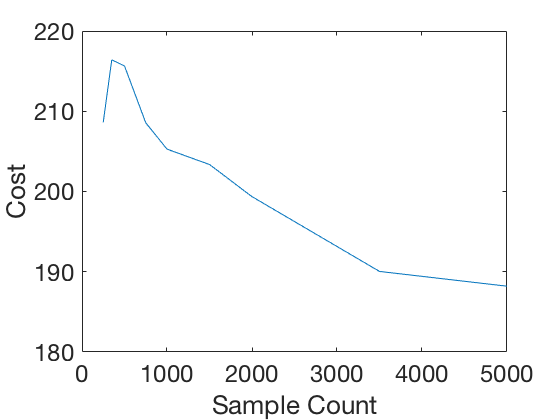}
        \caption{Cost}
        \label{fig:humanoidCost}
    \end{subfigure}
    \begin{subfigure}{0.22\textwidth}
        \includegraphics[width=\textwidth]{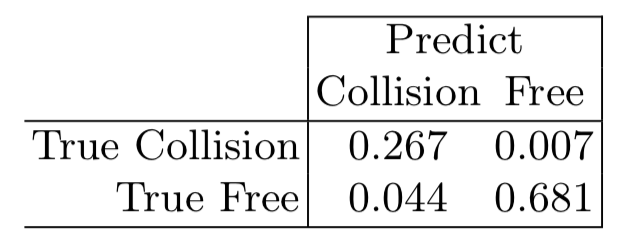}
        \caption{Collision predictions}
        \label{fig:humanoidConfusion}
    \end{subfigure}
   \caption{Convergence curves and collision confusion table for the humanoid problem.}
\label{fig:humanoidConvergence}
\end{figure}

For a single problem, \algtree for the humanoid (projected into two and three dimensions) is shown in Fig. \ref{fig:humanoidRRT}. 
The latent search is able to remain near the learned manifold, and the collision checker has effectively encapsulated the obstacles.
The final trajectory is shown in Fig. \ref{fig:humanoidRun} with the control input (the target location) set as a half second ahead on the planned trajectory.
A video is available at \url{goo.gl/wrm9JR}. 
Fig. \ref{fig:humanoidConvergence} details \algtree's performance in terms of success rate, convergence, and collision network confusion.
After only a few hundreds of samples, the algorithm is able to solve 100\% of the planning problems.
The trajectory cost again shows convergence as the number of samples increases.
The collision checking network performs well, classifying nearly 95\% correct, with only a 0.7\% false positive rate.
With 1000 samples, the algorithm takes on average 15 seconds.
Note we do not provide comparisons to planning in the full state space as traditional planning approaches with the 50 dimensional, dynamical system are intractable (unlike the visual planning problem where an exact lower-dimensional state was available).

\section{Conclusions}\label{sec:conclusions}

We have presented the Latent Sampling-based Motion Planning (\algname) methodology, which leverages the effectiveness of local, latent representations for robotic systems with techniques from sampling-based motion planning (SBMP) to compute motion plans for high-dimensional, complex systems.
In particular, this latent space is learned through an autoencoder, a dynamics network, and a separate collision checking network, each of which enforces the main algorithmic primitives of SBMP on the latent space (sampling, local connections, and collision checking).
Given this latent space, we use the Learned Latent RRT (\algtree) algorithm to globally explore the latent space and compute motion plans directly in it.
Through two experiments, one planning within visual space and one planning with a humanoid robot, we demonstrate the methodology's overall generality and ability to generalize to new environments.

This work leaves several avenues for future work. 
Specific to \algname, we first plan to investigate how much data is required to learn the necessary representation.
Second, we plan to investigate learning more restricted classes of dynamics for which a steering function is available, e.g., locally-linear dynamics. 
This would allow more optimal latent space exploration techniques.
Third, we plan to investigate using unsupervised learning to learn the collision checking network. 
Fourth, we plan to investigate planning problems in which the topology of the latent space is dependent on the environment and obstacle set specific to a planning problem, e.g., if stairs are involved for a humanoid.
Fifth, we plan to use a similar methodology within a task and motion planning framework by state augmentation in the full state space.
Finally, we plan to investigate conditions under which theoretical guarantees for \algtree can be derived.
Beyond \algname, the approach of learning plannable latent spaces through enforced algorithmic primitives can be quite powerful. 
We believe a similar methodology could be used to learn a space directly for trajectory optimization, or instead for lower-dimensional problems, combinatorial motion planning. 

\section*{Acknowledgments}
The authors wish to thank Edward Schmerling and Anirudha Majumdar for their helpful discussions on this work.

\renewcommand{\baselinestretch}{.94}
\bibliographystyle{IEEEtran-short}
\bibliography{../../../../bib/main,../../../../bib/ASL_papers,cite}

\end{document}